\documentclass[letterpaper, 10 pt, conference]{ieeeconf}
\IEEEoverridecommandlockouts
\usepackage{graphicx}
\usepackage{amsmath}
\usepackage{amsfonts}
\usepackage{color}
\usepackage{url}
\usepackage{amssymb}
\usepackage{cite}
\usepackage[linesnumbered,ruled]{algorithm2e}

\usepackage{amsthm}
\usepackage{subfigure}
\usepackage{caption}
\usepackage{multirow}
\usepackage{mathtools}

\DeclareMathOperator*{\argmax}{arg\,max}

\let\oldemptyset\emptyset
\let\emptyset\varnothing
\newcommand\inv[1]{#1\raisebox{1.15ex}{$\scriptscriptstyle-\!1$}}

\newtheorem*{prob}{Problem}

\definecolor{amber}{rgb}{1.0, 0.75, 0.0}
\newcommand{\changed}[1]{{\color{black}{#1}}}
\newcommand{\eg}{\textit{e.g.}}
\newcommand{\ie}{\textit{i.e.}}
\newcommand{\etal}{\textit{et al}.}

\title{\LARGE \bf
Environmental Hotspot Identification \changed{in Limited Time} with a UAV Equipped with a Downward-Facing Camera
}

\author{Yoonchang Sung, Deeksha Dixit, and Pratap Tokekar
\thanks{*This work was supported by the National Science Foundation under Grant No. 1637915.
(Corresponding author: Yoonchang Sung.)}%
\thanks{Y. Sung is with CSAIL, MIT, Cambridge, MA 02139, USA. \texttt{\small yooncs8@mit.edu}. D. Dixit and P. Tokekar are with the Department of Computer Science, University of Maryland, College Park, MD 20742, USA. {\tt\small \{deeksha,tokekar\}@umd.edu}. The part of the work was completed when all authors were with the Department of Electrical
and Computer Engineering, Virginia Tech, Blacksburg, VA 24061, USA.}
}

\begin{document}

\captionsetup{font=footnotesize}
\captionsetup[sub]{font=footnotesize}

\maketitle
\thispagestyle{empty}
\pagestyle{empty}

\begin{abstract}
\changed{Our work is motivated by environmental monitoring tasks, where finding the global maxima (\ie, hotspot) of a spatially varying field is crucial. We investigate the problem of identifying the hotspot for fields that can be sensed using an Unmanned Aerial Vehicle (UAV) equipped with a downward-facing camera. The UAV has a limited time budget which it can use for learning the unknown field and identifying the hotspot. Our contribution is to show how this problem can be formulated as a novel multi-fidelity variant of the Gaussian Process (GP) multi-armed bandit problem. The novelty is two-fold: (i) unlike standard multi-armed bandit settings, the rewards of the arms are correlated with each other; and (ii) unlike standard GP regression, the measurements in our problem are images (\ie, vector measurements) whose quality depends on the altitude of the UAV. We present a strategy for finding the sequence of UAV sensing locations and empirically compare it with several baselines. Experimental results using images gathered onboard a UAV are also presented and the scalability of the proposed methodology is assessed in a large-scale simulated environment in Gazebo.}
\end{abstract}


\section{Introduction}\label{sec:intro}


Robots are predominantly tasked with monitoring unknown environments but \changed{sometimes} their limited sensing capabilities restrict them from observing the entire environment at once. It is thus of importance to actively explore the environment and learn the underlying characteristics of the environment. 
Given the limited resources (\eg, operation time and fuel), the robot must carefully choose its actions
to better estimate and predict states of the environment.


\begin{figure}[thpb]
\centering
\includegraphics[width=0.60\columnwidth]{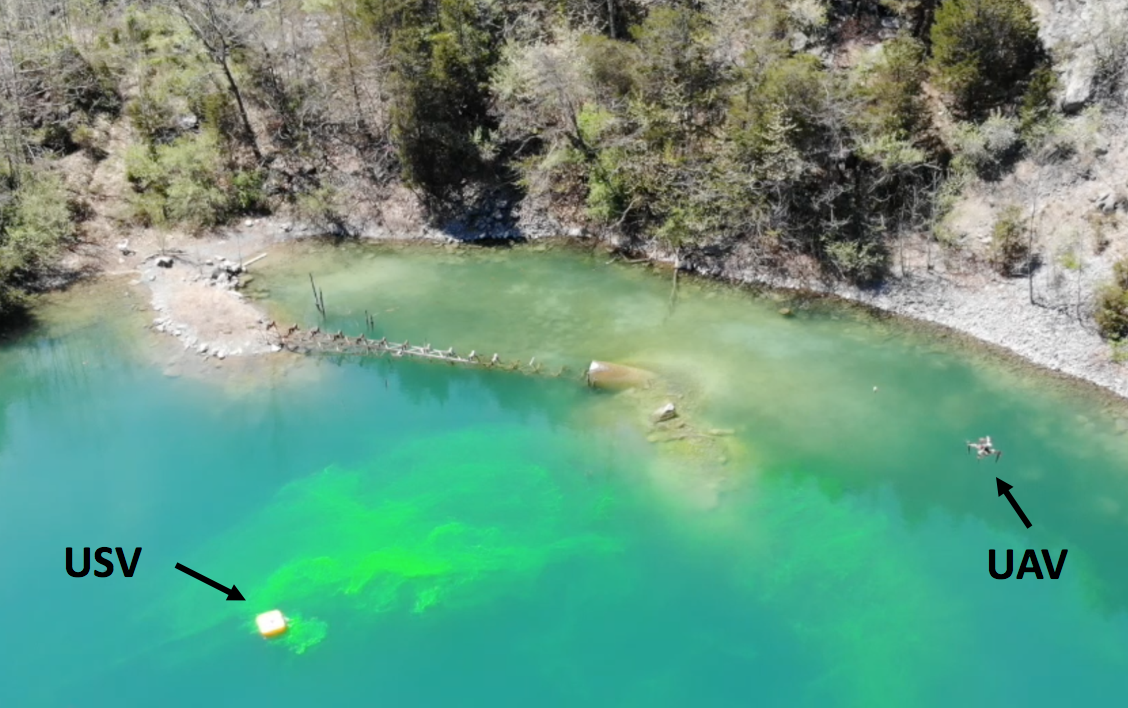}
\caption{A UAV exploring the environment to search for the plume in a lake in Blacksburg, Virginia~\cite{sung2019competitive}.}
\label{fig:motivation}
\end{figure}
This work is motivated by one such problem of monitoring hazardous plumes of pollutants released in water bodies, such as oil spills. The overarching project~\cite{nsf} is a collaboration with microbiologists interested in studying the transport of aerosolized pollutants from water bodies~\cite{powers2018tracking}. Figure~\ref{fig:motivation} demonstrates our team of an Unmanned Aerial Vehicle (UAV) and Unmanned Surface Vehicle (USV) cooperatively monitoring a lake. The UAV observes regions in the lake with a downward-facing camera using which it can map out the concentration of the (visible) hazardous agent. To analyze the characteristics of toxic particulates, just mapping it with UAVs is not enough. Instead, we need physical samples that can be analyzed ex situ. The UAV can direct the USV to the location with the highest concentration. The USV can then collect the physical specimen at that location. The UAV acts as an explorer whereas the USV acts as a sampler. Our focus is on planning strategies for the UAV to find the location with the highest concentration in the limited battery life.

While the aforementioned application motivates our work, the problem we study is general enough to apply to many settings where UAVs with downward-facing cameras can be used for finding \changed{hotspots (\ie, location of the global maxima) of unknown, spatially varying function}. There exist many practical applications in the environmental monitoring literature, such as precision agriculture~\cite{tokekar2016sensor},  wildlife habitat monitoring~\cite{cliff2015online}, plume tracking~\cite{sung2019competitive}, where such a problem might arise.

\changed{Since the measurements are noisy, the accuracy of the estimate can be improved by increasing the number of measurements from a sensing location.}
However, spending too much time at one location is not beneficial since the UAV has a limited budget. This is the exploration-exploitation dilemma studied under the Multi-Armed Bandit (MAB) setting (we introduce several variants of MAB in Section~\ref{sec:related}).

Our problem poses two major challenges that cannot be handled by existing MAB approaches. First, as the UAV can change its flight altitude (or, equivalently, the size of the camera footprint), we need different ways of evaluating the sensing performance at different altitudes.  
Second, there is no direct correspondence between the camera image and the measurements. Moreover, we need to define how good a particular sensing location is with respect to other locations.

Our approach is based on the Gaussian Process Upper Confidence Bound (GP-UCB) algorithm proposed by Srinivas~\etal~\cite{srinivas2010gaussian} that adopts the Gaussian Process (GP) to resolve spatially-correlated sensing locations. Our algorithm extends this to deal with the above challenges to be applicable in a  3D environment. 
In particular, we investigate the \emph{multi-fidelity} aspect in GP to handle varying sensing performance at different altitudes.
We then propose several heuristic planning strategies to qualitatively evaluate the proposed algorithm. We further exploit sparse approximation studied in~\cite{quinonero2005unifying,titsias2009variational} in our framework to overcome the cubic complexity of the GP regression.
The sparse approximation methods are able to handle the large observation spaces induced by the image measurements.

The contributions of this paper are as follows:
(1) We propose 
the Multi-Fidelity GP-UCB (MF-GP-UCB) algorithm
for environmental monitoring in a 3D environment with a camera sensor that has noise proportional to the altitude.
(2) We empirically show that the proposed strategies outperform numerous baselines through extensive simulations.
(3) We demonstrate how our algorithms can be deployed on the actual hardware using real-world data \changed{and in large-scale simulated environments in Gazebo}.

Our validation is based on real-world data we gathered from the field using a UAV \changed{and a physics simulation (\ie, Gazebo) for large-scale experiments}. We empirically show the performance of several planning algorithms through simulations. These results demonstrate the effectiveness of the Conditional Predictive Variance (CPV) that we propose in the MAB framework which reflects various flight altitudes of the UAV. In particular, we show that using the CPV resulted in approximately $10\%$ improvement over the standard algorithms. We also find that using sparse GP techniques result in negligible performance drop but orders of magnitude lower computational times.



\section{Related Work}\label{sec:related}


Multiple algorithms have been proposed for environmental monitoring. In this section, we briefly introduce some of the recent works. For survey results, see~\cite{dunbabin2012robots}.

\changed{
Although various UAV-based field estimation methods have been proposed (\eg, \cite{lan2016rapidly}), our focus is to find global maxima or hotspot in unknown environments within a limited budget.
Hotspot identification has been widely studied in the robotics community in various forms such as probabilistic classification of hotspots in a GP-based environment~\cite{low2012decentralized}, source localization in a field of non-continuous, non-smooth gradients~\cite{hajieghrary2016multi}, and via a model-based approach to plume monitoring~\cite{silic2019field}. 
}

Offline algorithms compute a trajectory for the robot before the operation. Such coverage planning algorithms~\cite{galceran2013survey} need to know the size and shape of a given environment a priori. Either a predefined trajectory (\eg, boustrophedon path~\cite{choset2000coverage}) or tree search algorithms, such as depth-first search, can be utilized to completely cover the environment. 

If only partial information about the environment is given, the robot must be able to adaptively cope with unexpected situations and uncertain environments online. Many decision-making challenges arise from this context. In environmental monitoring, the robot has to decide what samples to collect if samples are available only for a specific time period~\cite{flaspohler2018near,manjanna2018heterogeneous}, or where to take measurements when constrained to a limited budget~\cite{sung2019competitive}.

Information gathering is also an important task for understanding the environment. In particular, \emph{variance reduction}~\cite{ma2018multi,hollinger2014sampling} and \emph{mutual information maximization}~\cite{binney2013optimizing} are two popular information metrics. However, these approaches are not suitable for identifying hotspots in the environment as they are interested in exploring the environment rather than exploiting the current knowledge of learned information. 

Bayesian optimization~\cite{snoek2012practical} addresses this issue by balancing between exploration and exploitation. Previous works~\cite{slivkins2019introduction} on MAB problems proposed algorithms to identify a near-optimal arm within a given budget. They also proved some guarantees on the regret, although costs of traveling between arms (\ie, switching cost) were not considered. \changed{Regret is a measure of suboptimality of the reward obtained by the robot given as the difference between the optimal reward and the actual reward obtained by the robot.}

There exist variants of MAB considering switching costs, which can be useful in robotics because traveling cost is a bottleneck for the budget. Reverdy~\etal~\cite{reverdy2014modeling} employed a block allocation scheme that restrains the robot from switching to other arms frequently. Guha and Munagala~\cite{guha2009multi} developed an algorithm related to the orienteering problem~\cite{vansteenwegen2011orienteering} to minimize traveling costs while maximizing collected rewards. However, their algorithm cannot handle spatially correlated arms. Audibert and Bubeck~\cite{audibert2010best} studied the problem of finding the best arm at a given confidence level by the minimal number of arm pulls. Their objective, however, is to optimize the unknown budget whereas in our problem the budget is given as an input. Those works~\cite{vansteenwegen2011orienteering,audibert2010best} used a terminal regret metric for evaluating their algorithms. \changed{We use terminal regret as our evaluation metric. This is particularly suitable in our setting since the maxima located will be where the USV will obtain a physical sample from. Hence, the final output at the end of the learning process is of interest, rather than cumulative regret.}

To address varying degrees of accuracy of measurements in GP-UCB, multi-fidelity GP-UCB has recently been proposed~\cite{kandasamy2019multi,song2019general}. Compared to these works, in our setting, we obtain multi-fidelity measurements but we also have switching costs and the underlying geometry where the Field-Of-View (FOV) increases at lower fidelity (\ie, higher altitude) levels. A similar setting is studied in~\cite{wei2020expedited} but their objective is multi-target search.

We implemented information-gathering approaches and a MAB variant with switching costs~\cite{reverdy2014modeling} introduced here for comparison analysis in Section~\ref{subsec:comparison}.



\section{Problem Description}\label{sec:prob}



Let $\mathcal{E}\subseteq\mathbb{R}^2$ represent the 2D environment that contains the contamination and $\textbf{x}\in\mathcal{E}$ be a point within the environment. The intensity of the contamination varies within $\mathcal{E}$ but we assume that the intensity does not change during the course of the robot deployment.

The goal of this paper is to find a point in the 2D environment that has the highest intensity of contamination, denoted by $\textbf{x}^{OPT}\in\mathcal{E}$. A UAV is deployed to explore the environment, constrained by the limited time allowed to search for this particular point. The total time allocated is denoted by budget $B\in\mathbb{R}_{>0}$. The time spent is computed as the combination of \emph{sensing time} $T_S$ and \emph{traveling time} function $T_T(\textbf{x},\textbf{x}^\prime)$. We assume that sensing time $T_S$ is the same for any points $\textbf{x}$ regardless of the UAV position. While $T_S$ is a fixed constant value, $T_T(\textbf{x},\textbf{x}^\prime)$ is a function that takes as input two locations of the UAV and outputs the time 
for traveling from $\textbf{x}$ to $\textbf{x}^\prime$, assuming that the UAV moves with unit speed. The highest contamination point inferred by the UAV at the end of the budget $B$ will then be visited by the USV to physically collect a sample to be analyzed.


The UAV is mounted with a downward-facing camera to observe the contamination in the environment. We denote the state of the UAV at the $k$-th sensing location by $\textbf{v}(k)\in\mathbb{R}^3$. The state is composed of the x, y, and z coordinates of the UAV. At every sensing location, the UAV collects an image of the environment from a limited FOV camera sensor. 

Let $f(\cdot):\mathbb{R}^2\rightarrow\mathbb{R}_{\ge 0}$ be the true unknown intensity function. Each image gives a noisy observation of the true intensity function. We assume that the UAV can use image processing and other estimation techniques to fuse the data obtained from all images and form an estimate of $f(\textbf{x})$ at all locations $\textbf{x}$ in the environment. Based on its own estimate of the point of highest intensity, the UAV must determine a point $\textbf{x}^{ALG}\in\mathcal{E}$ that it believes to be the global maxima of $f$.
We wish to find strategies that will minimize the regret $f(\textbf{x}^{OPT}) - f(\textbf{x}^{ALG})$ at the terminal time $B$.
Since the UAV has a limited budget $B$, the number of images that can be collected by the UAV is also limited. Thus, sensing locations must be carefully chosen as the information about the unknown intensity is revealed online.

Without learning how the unknown intensities are varied over the environment, the UAV cannot identify the \changed{hotspot}. Thus, the UAV explores unvisited regions to gather more information while carefully spending the assigned budget. \changed{To sum up, we propose the following problem.}

\begin{prob}(\textbf{Hotspot Identification})

Let $f(x)$ be the true intensity function and $\textbf{\emph{x}}^{OPT}$ be the location in $\mathcal{E}$ where $f(\cdot)$ achieves the global maxima. Given the starting position of the UAV, $\textbf{\emph{v}}(0)$, find a strategy that produces outputs: (1) a sequence of sensing locations, $\textbf{\emph{v}}(1),\textbf{\emph{v}}(2),...,\textbf{\emph{v}}(k)$; and (2) a point, $\textbf{\emph{x}}^{ALG}\in\mathcal{E}$, to minimize a terminal regret $f(\textbf{\emph{x}}^{OPT})-f(\textbf{\emph{x}}^{ALG})$ subject to the constraint that
$\sum_{k=1} \big(T_T\big(\textbf{\emph{v}}(k), \textbf{\emph{v}}(k+1)\big) + T_S\big) \le B$.


\end{prob}


\section{3D MF-GP-UCB Algorithm}\label{sec:alg}

In this section, we show how to model this problem as a novel multi-fidelity variant of the MAB problem. In a typical MAB, we are given a set of arms with initial unknown reward distributions. The goal is to find the arm with the highest expected reward.


\subsection{Arm Locations}\label{subsec:arm}


\begin{figure}[htb]
\centering
\subfigure[Arm placement.]{\includegraphics[width=0.42\columnwidth]{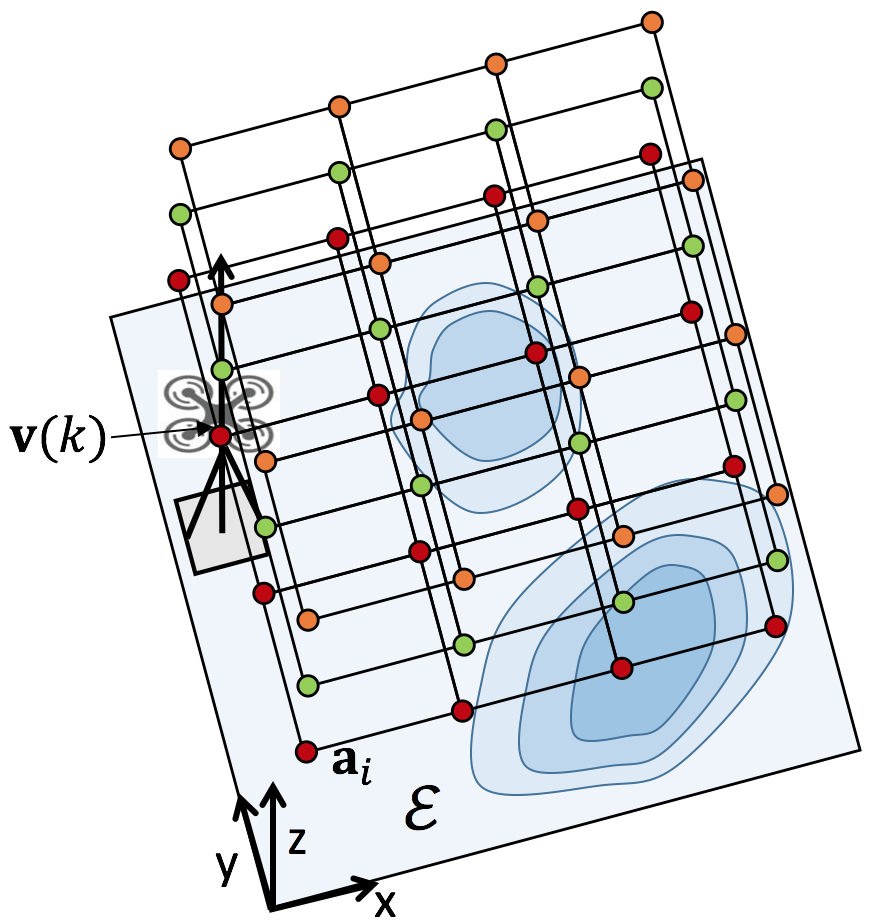}}
\subfigure[Image-to-measurement corresponden-ce.]{\includegraphics[width=0.56\columnwidth]{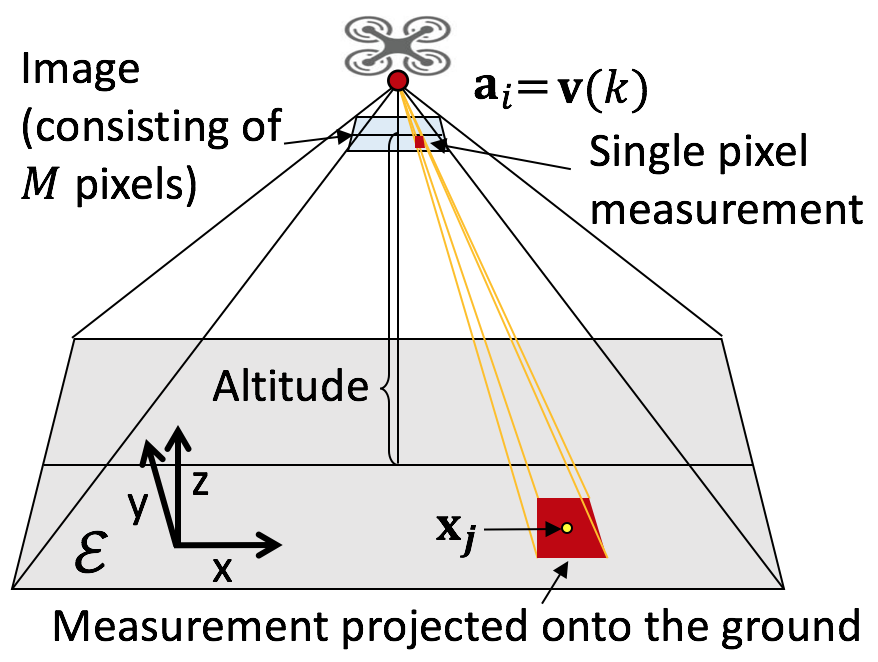}}
\caption{\changed{Illustration of our problem setting. In (a), arms (small circles) at three different altitudes (colored red, green, and orange for increasing altitude) are placed in a given 2D environment $\mathcal{E}$.
In (b), how the obtained image can be converted into measurements from an arm $i$ is described.}
}
\label{fig:env}
\end{figure}



We call sensing locations as \emph{arm locations} since that is the standard terminology in the MAB literature.
We create a 3D grid where each grid location is an arm (Figure~\ref{fig:env} (a)).
We denote the arm locations by $\textnormal{A}=\{\textbf{a}_1,...,\textbf{a}_N|\textbf{a}_i\in\mathbb{R}^3\}$ where $N$ is the total number of arm locations. We use $i$ to denote the index of an arbitrary arm.
We place the grid such that every point in the environment will be in the FOV of at least one arm at the lowest altitude.



\subsection{Sensor Model}\label{subsec:sensor}
At each arm location, the UAV obtains a camera image from the downward-facing camera. 
The size of the camera footprint is proportional to the altitude of the UAV, \ie,
the camera footprint will be larger from higher altitudes.


A single image measurement yields $M$ pixel measurements over $\mathcal{E}$. By using the camera projection equations with the known intrinsic, extrinsic parameters and known camera height, we can compute points in $\mathcal{E}$ that corresponds to $M$ measurements~\cite{hartley2003multiple}.
We will provide an example of this transformation in Section~\ref{subsec:real-world}.


\subsection{Reward}\label{subsec:reward}
We assume that there is a function that takes an image as input and produces a noisy estimate of the true intensity function $f(\cdot)$ for every pixel. The reward is a non-negative real-valued function of the concentration of the contamination in the area covered by that pixel. In Section~\ref{sec:sim}, we describe the reward function we use. 

We denote the reward value of the $j$-th measurement by $y_j\in\mathbb{R}_{\ge 0}$. The reward value can be computed from noisy estimates of the true intensity function $f(\cdot)$:
\begin{equation}~\label{eqn:reward}
y_j=f(\textbf{x}_j)+\epsilon,
\end{equation}
where $\epsilon\sim \mathcal{N}(0,\sigma^{2}_N)$ is additive i.i.d. Gaussian noise. 
The point $\textbf{x}_j$ corresponds to the $j$-th measurement position. 
Note again that larger the value of $f$ higher the contamination.



At each sensing location, the UAV obtains $M$ measurements where one pixel corresponds to one measurement. This measurement is the noisy version of the true intensity function of the location at the center of the footprint of the pixel on the ground (see Figure~\ref{fig:env} (b)). Therefore, after taking $k$ images, we have a collection of observed intensity values of $kM$ measurements, which we denote by $\textnormal{Y}\subseteq\mathbb{R}^{kM\times 1}$. The corresponding $kM$ measurement locations are denoted by $\textnormal{X}$. Note that these $kM$ locations denoted by $\textnormal{X}\subseteq\mathbb{R}^{kM\times 2}$ are not the same as the $k$ sensing locations (\ie, arm locations) where the $k$ images are obtained from.


Notice that the footprint of a pixel is not a point but an area, as the red quadrangle shown in Figure~\ref{fig:env} (b). The size of the footprint of a pixel gets larger as the altitude of the UAV becomes higher. The larger footprint size would not give an accurate reward value from a particular point $\textbf{x}$ compared to the smaller footprint size. \changed{The measurement noise variance $\sigma^{2}_N$ is proportional to the altitude of the UAV. Instead of using a fixed $\sigma^{2}_N$, we use $\sigma^{2}_N(\textbf{a}_i)$ that can vary depending on the arm altitude.}
We assume that the proportion of $\sigma^{2}_N(\textbf{a}_i)$ with respect to the arm altitude is linear.


\subsection{Multi-Fidelity Gaussian Process}\label{subsec:gp}

In this subsection, we discuss how the UAV proceeds with obtained measurements and maintains its belief of the contamination density distribution in the environment.

We use GP to represent the belief over time (\ie, $f\sim \mathcal{GP}$). Unlike conventional GP, measurements we obtain have various noise levels based on which altitude the measurements were observed. This is related with the noise variance $\sigma^2_N(\textbf{a}_i)$ at an arm $i$ (one of hyperparameters of the GP~\cite{rasmussen2003gaussian}), as explained in Section~\ref{subsec:reward}. Therefore, we adapt GP to our case that can take into account various noise levels.

When we learn the hyperparameters (length-scale, signal variance, and noise variance) offline by using the log marginal likelihood, $\sigma^2_N(\textbf{a}_i)$ is larger for data obtained from higher altitude than the one from lower altitude, \eg, $\sigma^2_N(\textbf{a}_i)>\sigma^2_N(\textbf{a}_j)$ if the altitude of arm $i$ is higher than that of arm $j$. We define the noise variance matrix given by:
\begin{equation}~\label{eqn:variance}
Q(\textnormal{X})=
\begin{bmatrix} 
\sigma^2_N(\textbf{a}_i) &  & \text{0} \\
 & \ddots & \\
\text{0} & & \sigma^2_N(\textbf{a}_i) 
\end{bmatrix},
\end{equation}
which is a diagonal matrix. Each diagonal element of $Q(\textnormal{X})$, $\sigma^2_N(\textbf{a}_i)$, has a unique value based on the altitude of the $i$-th arm, allowing diagonal elements of $Q(\textnormal{X})$ to have various values, but we set $\sigma^2_N(\textbf{a}_i)$ for arms at the same altitude to have the same value. The dimension of $Q(\textnormal{X})$ is proportional to the total number of measurements accumulated during the flight, \ie, $Q(\textnormal{X})\subseteq\mathbb{R}^{kM\times kM}$ at sensing location $k$.

Given a 2D grid of the environment, let $L_i$ be the number of grid cells that fall in the camera footprint of arm $i$ which we call test points. Let $I_i$ be the index set containing indices of test points 
observed from an arm $i$. 
We denote the set of $L_i$ test points by $\textnormal{X}^*_i=\{\textbf{x}^*_{I_i(1)},...,\textbf{x}^*_{I_i(L_i)}\}\subset \mathcal{E}$. Alternatively, $\textnormal{X}^*_i=\textbf{x}^*_{I_i}$ for an arm $i$. As we have $N$ arms, the total number of all test points from all arms becomes $L=\sum_{i=1}^NL_i$. The set of all test points from all arms is $\textnormal{X}^*=\cup_{i=1}^N\textnormal{X}^*_i=\{\textbf{x}^*_1,...,\textbf{x}^*_L|\textbf{x}^*_l\in\mathcal{E}\}$. 
The prior of the GP then becomes: 
\begin{equation}~\label{eqn:prior}
\begin{bmatrix} 
\textnormal{Y} \\
f_* \\
\end{bmatrix}
\sim\mathcal{N}
\begin{pmatrix} 
\multicolumn{2}{c}{\textbf{0},} & 
\begin{bmatrix} 
\mathcal{K}(\textnormal{X}, \textnormal{X})+Q(\textnormal{X}) & \mathcal{K}(\textnormal{X}, \textnormal{X}^*) \\
\mathcal{K}(\textnormal{X}^*, \textnormal{X}) & \mathcal{K}(\textnormal{X}^*, \textnormal{X}^*) \\
\end{bmatrix}
\end{pmatrix},
\end{equation}
where $\mathcal{K}(\cdot, \cdot)$ is a covariance function (or kernel). In this work, we use the squared exponential covariance function~\cite{rasmussen2003gaussian}. The predictive mean $\mu$ and covariance $P$ are:


\begin{align}
&\mu=\mathcal{K}(\textnormal{X}^*, \textnormal{X})\inv{\big(\mathcal{K}(\textnormal{X}, \textnormal{X})+Q(\textnormal{X})\big)}\textnormal{Y},~\label{eqn:pred_mean}\\
&P=\mathcal{K}(\textnormal{X}^*, \textnormal{X}^*)-\mathcal{K}(\textnormal{X}^*, \textnormal{X})\inv{\big(\mathcal{K}(\textnormal{X}, \textnormal{X})+Q(\textnormal{X})\big)}\mathcal{K}(\textnormal{X}, \textnormal{X}^*).~\label{eqn:pred_cov}
\end{align}

The inversion of a high-dimensional matrix in Equations (\ref{eqn:pred_mean}) and (\ref{eqn:pred_cov}) makes the algorithm infeasible to run in real-time. This is a well-known difficulty in using GP and in our case the size of the inversion matrix scales as $\mathcal{O}(k^3M^3)$. To overcome the computational bottleneck, various techniques have been proposed to develop a sparse GP approximation~\cite{quinonero2005unifying,titsias2009variational}. These approaches generate a small set of $S$ inducing (or support) points to approximate the original GP such that $S\ll kM$, eventually reducing the time complexity to $\mathcal{O}(SkM)$. We empirically show the power of linear complexity owing to the sparse GP in Section~\ref{sec:sim}. By choosing a small enough $S$, one can adapt the algorithm for real-time applications. Specifically, we show an order of magnitude reduction in computational time without affecting the performance.


\subsection{Pseudo-code}\label{subsec:pseudo}

The pseudo-code in Algorithm~\ref{alg:gpucb} demonstrates all the steps of hotspot identification. The UAV starting from an initial position $\textbf{v}(0)=\textbf{a}_i(0)$ is given a budget $B$ and initializes GP with zero mean. At every sensing location $k$, the UAV updates the GP mean and variance functions (Lines 10 and 13) by using measurements collected up to and including the previous sensing locations (Line 20). To compute the objective function (Line 17), we calculate the average mean and the average variance (Lines 14 and 15) as the number of test points is not $1$. If the budget $B$ assigned to the UAV is exhausted (Line 2), the algorithm terminates and the UAV finds $\textbf{x}^{ALG}$ from learned GP for the USV to sample from this location.


\section{Planning Strategies}\label{sec:planning}
We present several heuristic planning strategies that can be used as a subroutine in Algorithm~\ref{alg:gpucb}. 
We explain the MF-GP-UCB algorithm first and then present the two changes we make to the algorithm for better accuracy, \ie, \emph{CPV} and \emph{dynamic window approach}.

We present the objective function (Line 17 of Algorithm~\ref{alg:gpucb}) that decides which arm the UAV should visit next. The weighted combination of mean and variance is the conventional functional form in MAB, however, $\beta$ values proposed by the UCB algorithms as well as the GP-UCB algorithm~\cite{srinivas2010gaussian} cannot directly be applied to our problem because of the challenges explained in Section~\ref{sec:intro}. Thus, we propose the following exponential functional form for $\beta$:
\begin{equation}~\label{eqn:objective}
\beta=\gamma e^{(\lambda k)},
\end{equation}
where $\gamma$ and $\lambda$ are hyperparameters that control the decreasing rate or the increasing rate of $\beta$. We tune them as well as the inherent GP hyperparameters offline using data collected in simulations.


GP-UCB we use is originated from Srinivas~\etal~\cite{srinivas2010gaussian} but extended to take into account our problem setting (\ie, MF-GP-UCB). Moreover, with Equations (\ref{eqn:pred_mean}) and (\ref{eqn:pred_cov}), MF-GP-UCB can handle various noise levels for collected measurements. Algorithm~\ref{alg:gpucb} without having Line 12 is this MF-GP-UCB version. Taking out Line 12 implies that the variance of test points is estimated without worrying about where the current measurements are observed. Since the flight altitude of the UAV affects the sensing credibility, we tackle this in the following strategy.

\setlength{\textfloatsep}{0pt}
\begin{algorithm}
\SetAlgoLined
\SetKwInOut{Input}{Input}
\SetKwInOut{Output}{Output}
\SetKwFunction{FMove}{MoveUAV}
\SetKwFunction{FMea}{GetMeasurements}
\SetKwFunction{FSensing}{SensorModel}
\SetKwFunction{Fterm}{TerminateExploration}
\Input{Initial position of the UAV: $\textbf{v}(0)=\textbf{a}_i(0)\in$A, GP prior: $\mu(0)=0$, $\sigma(0)$, $\mathcal{K}$, budget: $B$, arm set: $\textnormal{A}=\{\textbf{a}_1,...,\textbf{a}_i,...,\textbf{a}_N\}$, test set: $\textnormal{X}^*=\{\textbf{x}^*_1,...,\textbf{x}^*_l,...,\textbf{x}^*_L\}$, measurement set: $\textnormal{X}=\oldemptyset$, reward value set: $\textnormal{Y}=\oldemptyset$.
}

\For{\emph{$k=1,2,...$}}{
\If{\emph{$T_T\big(\textbf{v}(k),\textbf{v}(k\text{-}1)\big)+T_Sk> B$}}{
$\textbf{x}^{ALG}=\argmax_{\textbf{x}} \mu$.

$\small\Fterm\big(\textbf{x}^{ALG}\big)$.
}

\If{\emph{$k=1$}}{

Choose a nearest arm $i$, \ie, $\textbf{a}_i(1)=\textbf{v}(1)$.
}
\Else{

$\mu=\mathcal{K}(\textnormal{X}^*, \textnormal{X})\inv{\big(\mathcal{K}(\textnormal{X}, \textnormal{X})+Q(\textnormal{X})\big)}\textnormal{Y}$.


\For{\emph{$i=1,2,...,N$}}{
$\textnormal{X}^{\prime}=\textnormal{X}\cup\textbf{x}^*_{I_i}$.

$P|\textbf{a}_i=\mathcal{K}(\textbf{x}^*_{I_i}, \textbf{x}^*_{I_i})-\mathcal{K}(\textbf{x}^*_{I_i}, \textnormal{X}^{\prime})\inv{\big(\mathcal{K}(\textnormal{X}^{\prime}, \textnormal{X}^{\prime})+Q(\textnormal{X}^{\prime})\big)}\mathcal{K}(\textnormal{X}^{\prime}, \textbf{x}^*_{I_i})$ where $P\ni\sigma_l^2\ \forall l\in\textnormal{I}_i$.

$\overline{\mu}_i=\big(\sum_{l\in\textnormal{I}_i}\mu_l\big)\big/L_i$.

$\overline{\sigma}_i^2=\big(\sum_{l\in\textnormal{I}_i}\sigma_l^2\big)\big/L_i^2$.
}

$\textbf{a}_i(k)=\argmax_i\big\{\overline{\mu}_i+\beta \overline{\sigma}_i\big\}$.
}

$\textbf{a}_i(k)\leftarrow\small\FMove\big(\textbf{v}(k\text{-}1)\big)$.

$\big\{\textnormal{X},\textnormal{Y}\big\}\leftarrow\big\{\textnormal{X},\textnormal{Y}\big\}\cup\small\FMea\big(\textbf{a}_i(k)\big)$.
}
\caption{3D MF-GP-UCB with CPV}
\label{alg:gpucb}
\end{algorithm}



Although the predictive covariance $P$ in Equation (\ref{eqn:pred_cov}) considers various noise levels for the training points (\ie, accumulated measurements), this aspect is not addressed for predicting the covariance at test points in Equation (\ref{eqn:pred_cov}). That is, we predict the covariance at a test point as if we were to observe the test point from the altitude of the corresponding arm. We achieve this by adopting CPV, \ie, $P|\textnormal{A}$. 

We decompose the CPV $P|\textnormal{A}$ into $N$ CPVs, \ie, $P|\textbf{a}_i$ for each arm $i$. Then, the test points and the training points become $\textbf{x}^*_{I_i}$ and 
$\textnormal{X}^{\prime}=\textnormal{X}\cup\textbf{x}^*_{I_i}$, respectively, for an arm $i$. The decomposed CPV for the $i$-th arm can be computed as:
\begin{equation}~\label{eqn:conditional}
\begin{split}
P|\textbf{a}_i=&\mathcal{K}(\textbf{x}^*_{I_i}, \textbf{x}^*_{I_i})-\mathcal{K}(\textbf{x}^*_{I_i}, \textnormal{X}^{\prime})\times\\
&\inv{\big(\mathcal{K}(\textnormal{X}^{\prime}, \textnormal{X}^{\prime})+Q(\textnormal{X}^{\prime})\big)}\mathcal{K}(\textnormal{X}^{\prime}, \textbf{x}^*_{I_i}).
\end{split}
\end{equation}


We call this planning strategy as MF-GP-UCB with CPV. Similarly, we call the standard MF-GP-UCB as MF-GP-UCB with current variance.


The above strategies are not concerned with minimizing the total traveling cost of the UAV with respect to the traveling time function $T_T(\textbf{x},\textbf{x}^\prime)$ \changed{as the algorithm greedily chooses the best arm at each time}. If an algorithm keeps moving the UAV from one end of the environment to the other end, the UAV cannot gather much information and may output a low intensity point due to the limited budget $B$. 

\changed{We employ the dynamic window approach as a heuristic where the 3D window is defined centered on the current position of the UAV.} The UAV is only allowed to visit neighboring arms to observe at each sensing location. To do that in Algorithm~\ref{alg:gpucb}, the UAV considers neighboring arms (\ie, a subset of $A$) with respect to the current UAV position $\textbf{v}(k)$ in Lines 10-17 to decide which arm to visit next. 


\section{Simulations}\label{sec:sim}



We implemented Monte Carlo simulations using MATLAB to verify the performance of Algorithm~\ref{alg:gpucb}. We randomly generated $40$ environments ($20\times20$ square meters, having \changed{different distributions of} multiple local maxima but the same global maximum intensity value, \ie, $50$, so as to have the same $f(\textbf{x}^{OPT})$ for all environments) and ran $10$ instances for each environment. We set three altitudes for the UAV to fly at $10$, $40$, and $70$ meters. Each image consists of $9$ pixels and the camera footprint size is $1\times 1\ m^2$ at the lowest altitude, $4\times 4\ m^2$ at the middle altitude, and $7\times 7\ m^2$ at the highest altitude. We considered the total budget of $100$ seconds in all cases because this amount prevents the UAV from visiting all arms, otherwise resulting in a situation where careful decision-making is not required. Taking an image requires a total of $2$ seconds, \ie, $T_S = 2$ seconds. This time can be used to make a stop to take images without blur as well as to plan for subsequent actions.

We conducted an ablation study on planning strategies in Section~\ref{sec:planning} and comparison analysis with baseline algorithms. We then compared 3D planning with 2D planning where the UAV was allowed to fly at a fixed altitude. We also tested how the performance of algorithms varies depending on the amount of budget. Lastly, we studied the effect of sparsity to alleviate the computational complexity of the GP regression.

We define two terminal regrets as performance metrics: the \emph{point performance metric} to measure how close $f(\textbf{x}^{ALG})$ is to $f(\textbf{x}^{OPT})$ by $\frac{f(\textbf{x}^{ALG})}{f(\textbf{x}^{OPT})}\times 100\%$, and the \emph{arm performance metric} to measure how close $\sum_{\textbf{a}_i^{ALG}}f(\textbf{x})$ is to $\sum_{\textbf{a}_i^{OPT}}f(\textbf{x})$ by $\frac{\sum_{\textbf{a}_i^{ALG}}f(\textbf{x})}{\sum_{\textbf{a}_i^{OPT}}f(\textbf{x})}\times 100\%$. The point performance metric is relevant when a USV will go and collect a physical sample at $\textbf{x}^{OPT}$. The arm performance metric is relevant when, instead of a USV, a UAV will go and collect a final image measurement at the best arm location.


\subsection{Ablation Study}\label{subsec:ablation}
\changed{We evaluated eight heuristic planning algorithms where we consider the decreasing and increasing rates of $\beta$ as well as with and without the dynamic window approach.} 
From randomly generated environments, we tuned the hyperparameters ($\gamma$ and $\lambda$) of $\beta$ (Equation (\ref{eqn:objective})) offline through simulations for both the increasing rate and the decreasing rate. We used the following exponential functional forms:
\begin{itemize}
\item $\beta=1.5e^{-0.05k}$ for MF-GP-UCB with current variance of decreasing $\beta$. \item $\beta=10e^{-0.05k}$ for MF-GP-UCB with CPV of decreasing $\beta$. \item $\beta=-0.5e^{-0.05k}+0.5$ for MF-GP-UCB with current variance of increasing $\beta$. 
\item $\beta=-10e^{-0.05k}+10$ for MF-GP-UCB with CPV of increasing $\beta$.
\end{itemize}


From Table~\ref{tab:ablation}, we observe that taking into account the CPV consistently results in approximately $10\%$ improvement. This can be observed by comparing CV with CPV and DCV with DCPV. We notice similar improvements in all versions of the algorithm. Also, with the dynamic window approach, we observe approximately $25\%$ improvement on average.


\begin{table}[h]
  \centering
  \begin{tabular}{|c||c|c|}
    \hline
    \multirow{2}{*}{Algorithms} & \multicolumn{2}{c|}{Performance metrics ($\%$)} \\
    \cline{2-3}
     & Point & Arm \\
    \hline
    \hline
    MF-GP-UCB with CV ($\beta\text{-}\text{-}$) & $47.65\pm4.52$ & $46.31\pm4.51$ \\
    \hline
    MF-GP-UCB with CPV ($\beta\text{-}\text{-}$) & $57.40\pm4.83$ & $57.49\pm5.17$ \\
    \hline
    MF-GP-UCB with CV ($\beta\text{+}\text{+}$) & $49.39\pm4.81$ & $50.62\pm4.83$ \\
    \hline
    MF-GP-UCB with CPV ($\beta\text{+}\text{+}$) & $59.46\pm5.07$ & $58.49\pm5.14$ \\
    \hline
    MF-GP-UCB with DCV ($\beta\text{-}\text{-}$) & $62.94\pm4.89$ & $63.53\pm4.82$ \\
    \hline
    MF-GP-UCB with DCPV ($\beta\text{-}\text{-}$) & $71.28\pm4.82$ & $71.96\pm4.59$ \\
    \hline
    MF-GP-UCB with DCV ($\beta\text{+}\text{+}$) & $61.49\pm4.72$ & $62.19\pm4.59$ \\
    \hline
    MF-GP-UCB with DCPV ($\beta\text{+}\text{+}$) & $\mathbf{71.60\pm4.73}$ & $\mathbf{74.51\pm4.52}$ \\
    \hline
  \end{tabular}
  \caption{Ablation study on eight heuristic planning strategies. 
  The values represent the percents of how close the best-estimated value is to the true best value with the one-sigma error.
  Acronyms in this table are: CV (Current Variance), DCV (Dynamic window approach Current Variance), DCPV (Dynamic window approach CPV). $\beta\text{+}\text{+}$ and $\beta\text{-}\text{-}$ imply $\beta$ with the increasing rate and with the decreasing rate, respectively.}
  \label{tab:ablation}
\end{table}

\subsection{Comparison Analysis}\label{subsec:comparison}

\begin{table}[h]
  \centering
  \begin{tabular}{|c||c|c|}
    \hline
    \multirow{2}{*}{Algorithms} & \multicolumn{2}{c|}{Performance metrics ($\%$)} \\
    \cline{2-3}
     & Point & Arm \\
    \hline
    \hline
    MF-GP-UCB with DCPV ($\beta\text{+}\text{+}$) & $\mathbf{71.60\pm4.73}$ & $\mathbf{74.51\pm4.52}$ \\
    \hline
    Boustrophedon at high altitude & $35.86\pm3.16$ & $3.40\pm0.00$ \\
    \hline
    Boustrophedon at middle altitude & $34.81\pm2.74$ & $3.40\pm0.00$ \\
    \hline
    Boustrophedon at low altitude & $28.73\pm2.83$ & $3.40\pm0.00$ \\
    \hline
    Gradient ascent at high altitude & $14.61\pm7.55$ & $10.11\pm5.87$ \\
    \hline
    Gradient ascent at middle altitude & $14.03\pm10.52$ & $11.06\pm7.60$ \\
    \hline
    Variance reduction & $53.70\pm1.34$ & $52.49\pm1.31$ \\
    \hline
    Gradient ascent at low altitude & $13.69\pm9.56$ & $11.84\pm7.77$ \\
    \hline
    Mutual information maximization & $52.92\pm1.32$ & $48.60\pm1.22$ \\
    \hline
    Block UCL & $35.13\pm1.84$ & $11.91\pm0.45$ \\
    \hline
    MF-GP-UCB with D2DH ($\beta\text{-}\text{-}$) & $65.81\pm4.81$ & $69.58\pm4.52$ \\
    \hline
    MF-GP-UCB with D2DH ($\beta\text{+}\text{+}$) & $66.93\pm4.70$ & $72.12\pm4.50$ \\
    \hline
    MF-GP-UCB with D2DM ($\beta\text{-}\text{-}$) & $63.89\pm5.00$ & $69.08\pm3.93$ \\
    \hline
    MF-GP-UCB with D2DM ($\beta\text{+}\text{+}$) & $69.23\pm4.71$ & $71.22\pm4.12$ \\
    \hline
    MF-GP-UCB with D2DL ($\beta\text{-}\text{-}$) & $44.48\pm4.67$ & $45.88\pm4.60$ \\
    \hline
    MF-GP-UCB with D2DL ($\beta\text{+}\text{+}$) & $47.56\pm5.00$ & $48.31\pm4.91$ \\
    \hline
  \end{tabular}
  \caption{Comparison with eight heuristic planning strategies, four baseline algorithms, and 2D exploration algorithms. 
  We leave our algorithm with the best performance here as a comparison standard.
  Acronyms in this table are: D2DH (Dynamic window approach 2D exploration at the Highest altitude), D2DM (Dynamic window approach 2D exploration at the Middle altitude), and D2DL (Dynamic window approach 2D exploration at the Lowest altitude). 
  }
  \label{tab:all}
\end{table}


\subsubsection{Comparison with baseline algorithms}

We introduce four baseline algorithms that can also be applied to our problem: (1) the boustrophedon algorithm~\cite{choset2000coverage}, (2) gradient ascent method, (3) information-theoretic approach~\cite{krause2010sfo}, and (4) the block Upper Credible Limit (UCL) algorithm~\cite{reverdy2014modeling}. The boustrophedon algorithm is coverage planning where the entire environment is completely covered by the camera footprint of the UAV but information gathered online is not exploited. 
The gradient ascent method computes the gradient of the intensity function with respect to the current UAV position and the UAV moves in the direction of the positive gradient.
We implemented two variations of the information-theoretic approach: variance reduction and mutual information maximization, introduced in Section~\ref{sec:related}.
The block UCL algorithm is a Bayesian approach that addresses the exploration-exploitation dilemma. The unique feature of this algorithm is that it not only maximizes the expected reward obtained within a budget, but it also minimizes the number of switches to other arms, which is related to minimizing the traveling cost. 

We observe from Table~\ref{tab:all} that the proposed algorithm outperforms baseline algorithms in all cases. For the case of boustrophedon algorithms, the budget of $100$ was not sufficient to cover the entire environment and consequently ended up with poor performance. 
The gradient ascent methods at all altitudes performed the worst by getting trapped in local optima. Both variations of information methods outperformed other baselines, but not ours, as they explored the entire environment sufficiently while not concentrating on high-intensity regions. An environment having a peak on its boundary might be a bad case for information methods.
Due to the allocation scheme used by the block UCL, the algorithm requires sensing at an arm numerous times before moving to the next arm. This results in the exploration of only a small portion of the environment. 


\begin{figure}[thpb]
\centering
\includegraphics[width=0.66\columnwidth]{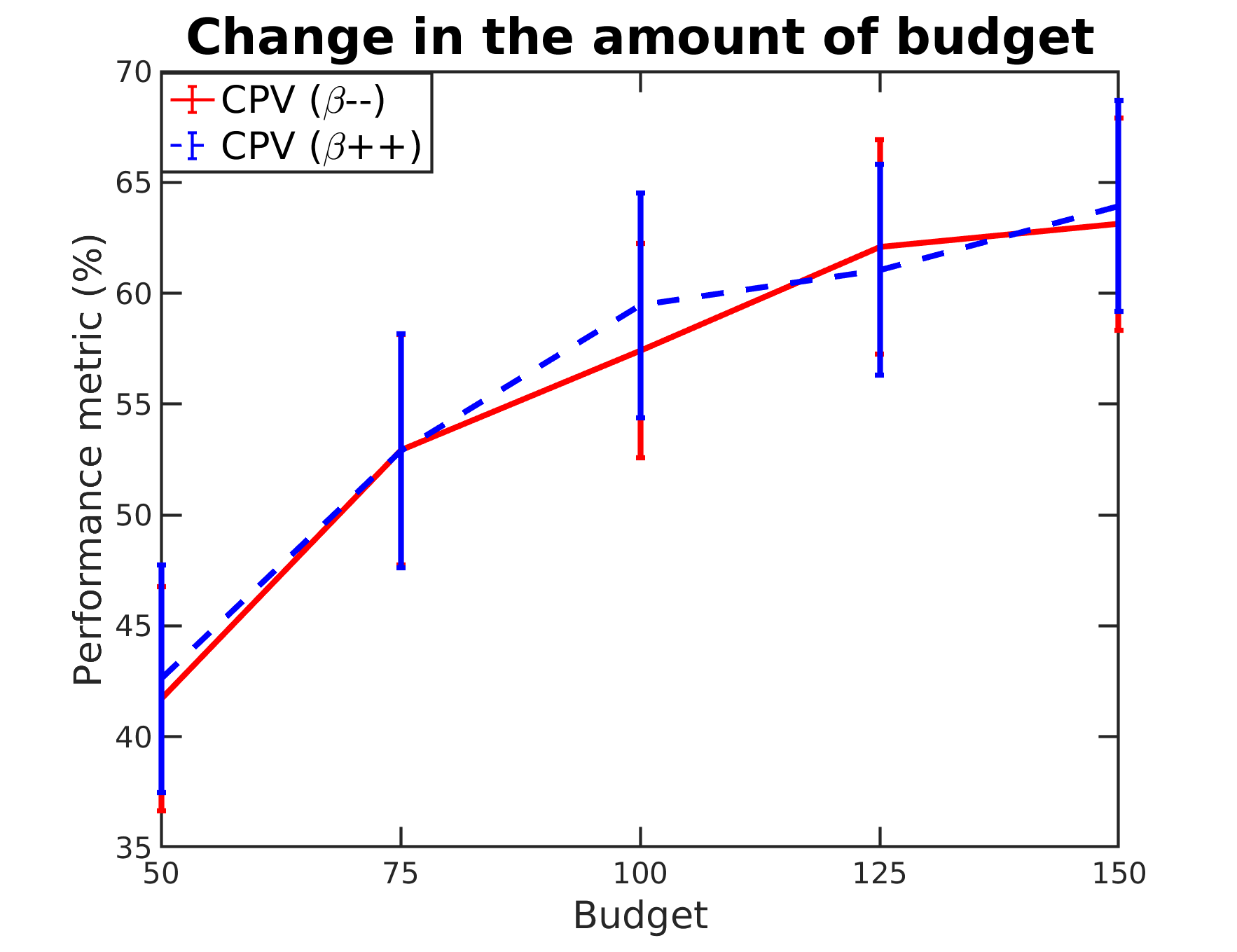}
\caption{Relationship between the budget and the performance metric for a point in case of MF-GP-UCB with CPV.}
\label{fig:budget}
\end{figure}


\subsubsection{Comparison with 2D planning}
We compared with 2D planning algorithms where we fixed the altitude at which the UAV can fly. It can be seen from Table~\ref{tab:all} that flying at a \changed{fixed} altitude is not beneficial in comparison with 3D exploration.



\subsubsection{Effect of the amount of budget}

We tested how the performance metric for a point varies with respect to the change in the amount of budget given to the UAV. As shown in Figure~\ref{fig:budget}, the performance metric increases logarithmically as the budget increases.

\subsubsection{Sparse GP evaluation}
\begin{figure}[htb]
\centering
\subfigure[The result of efficiency analysis.]{\includegraphics[width=0.49\columnwidth]{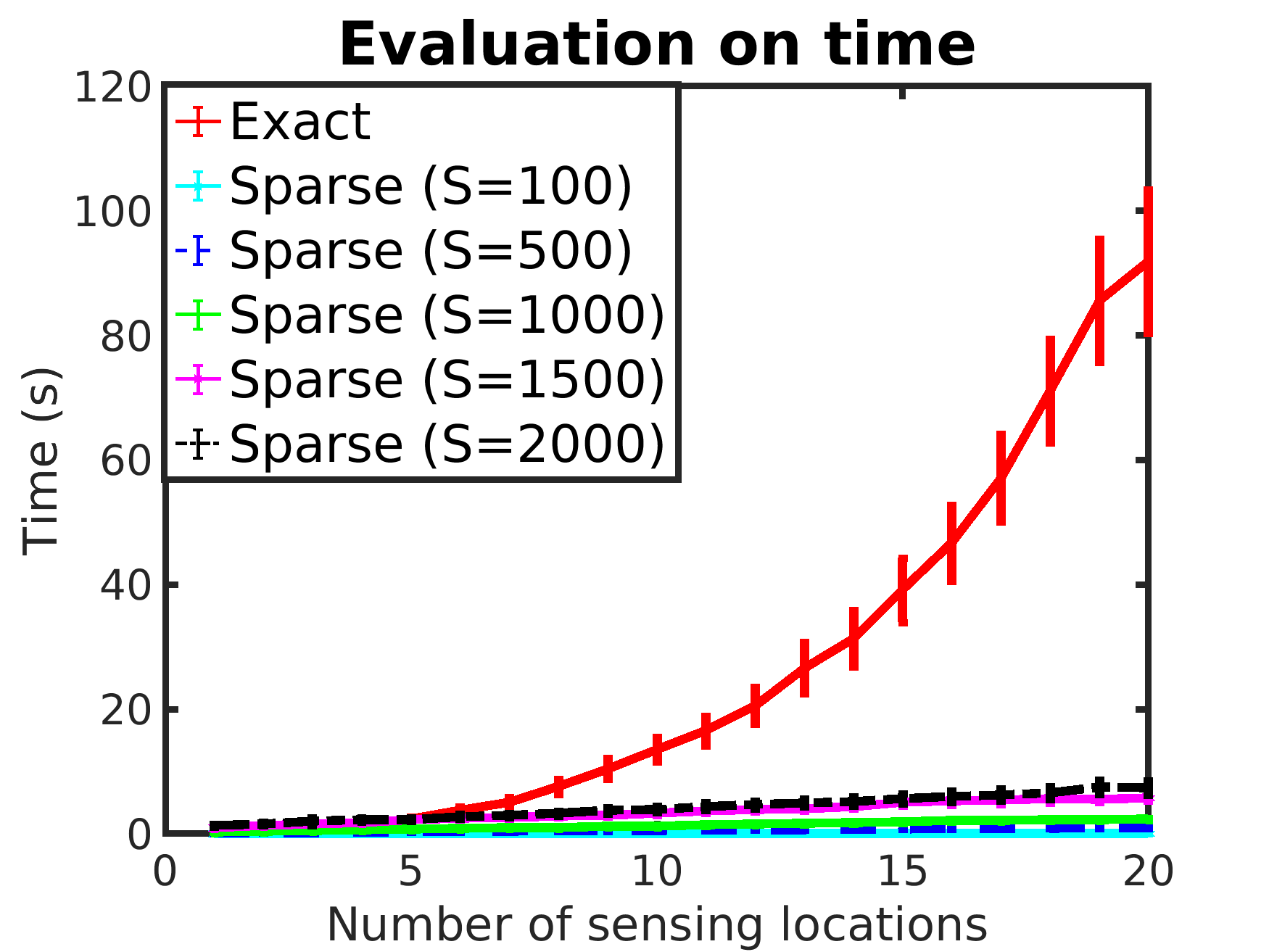}}
\subfigure[The result of performance analysis.]{\includegraphics[width=0.49\columnwidth]{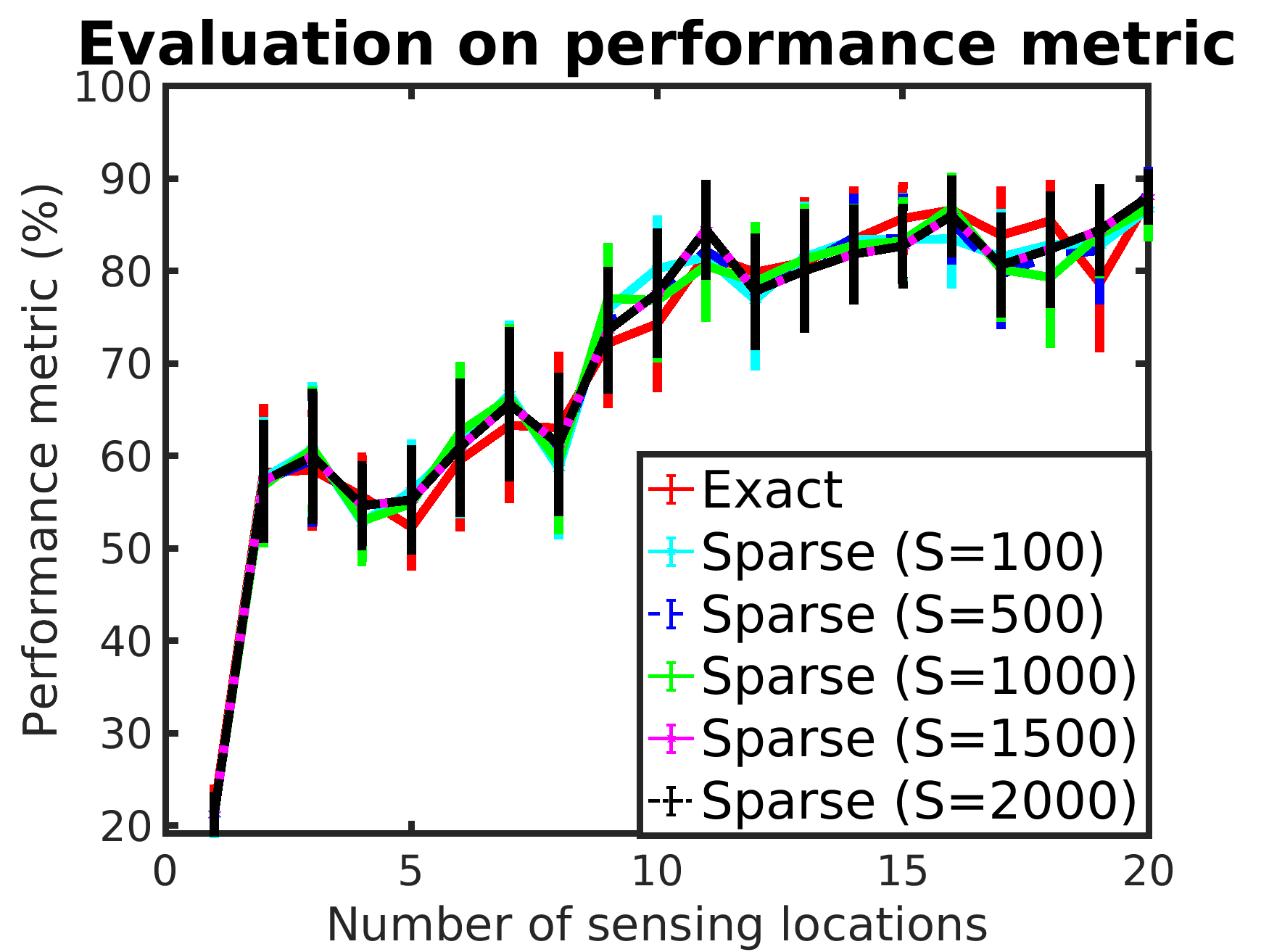}}
\caption{Comparison of time taken by the exact inference (\ie, the original GP) and the performance metric values with sparse approximations that have different numbers of inducing points $S$.}
\label{fig:sparsity}
\end{figure}

We compared the sparse GP approximation with the original GP to analyze how sparsity affects the performance of the proposed algorithm. We generated ten different environments to run algorithms $10$ times in each environment. At each sensing location, the robot received $646$ measurements. Plots in Figure~\ref{fig:sparsity} were obtained by computing the mean and standard deviation from all runs for each sensing location. We used different numbers of inducing points (\ie, $S$) for various levels of sparsity. 

As shown in Figure~\ref{fig:sparsity} (a), the time taken by the original GP increases exponentially due to the exact inference while the times taken by the sparse GPs grow linearly. However, we observe in Figure~\ref{fig:sparsity} (b) that there is little difference in performance metric values among the original GP and sparse approximations. From our analysis, the sparse GP yields orders of magnitude lower time than the original GP without any sacrifice of performance. Thus, we must take advantage of sparsity for real-time applications while introducing $S$ as an additional hyperparameter.


\subsection{Real-World Data}\label{subsec:real-world}

\begin{figure}[thpb]
\centering
\includegraphics[width=0.71\columnwidth]{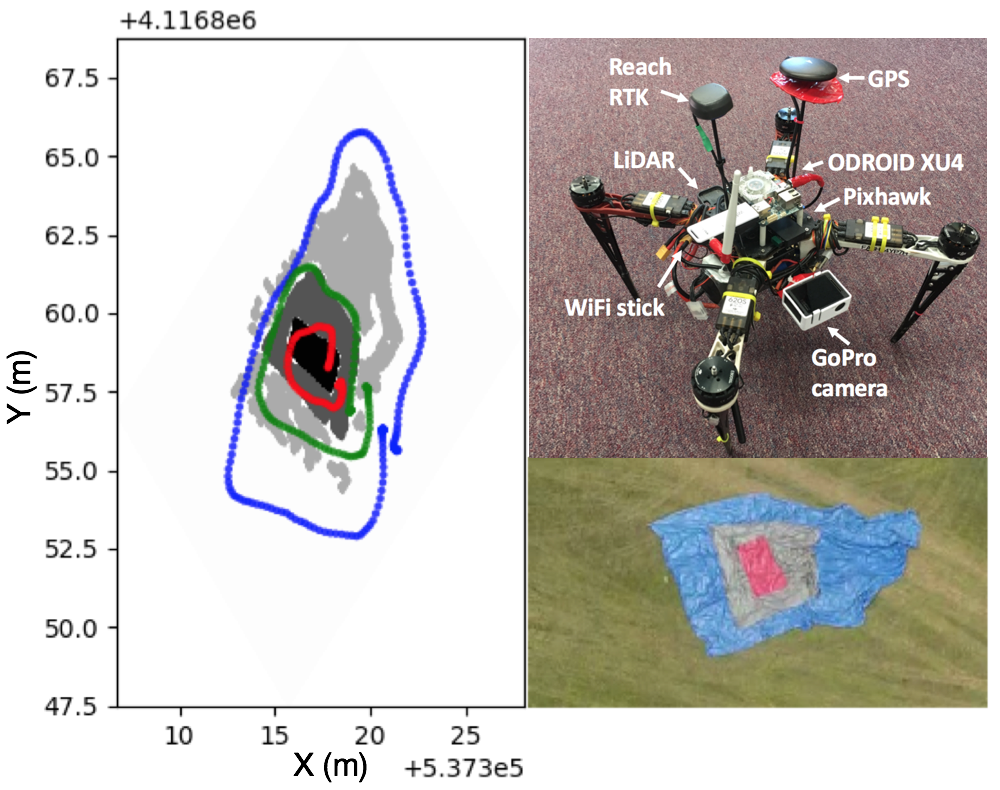}
\caption{Image on the top right shows the UAV platform. The bottom right image shows the tarps obtained by the UAV and used as a proxy of the regions with different intensities (red:3, gray:2, blue: 1). The image on the left shows the intensity for the image on the bottom right along with the ground truth boundaries. 
}
\label{fig:real_data}
\end{figure}

In our experimental setup, we used the UAV (Figure~\ref{fig:real_data}) with a single onboard-PC which has Linux 16.04 and ROS Kinetic~\cite{quigley2009ros} installed. It is equipped with a GPS sensor, a compass, and a downward-facing camera sensor (GoPro Hero4), which can communicate with the UAV over WiFi. \changed{The GPML MATLAB toolbox~\cite{rasmussen2003gaussian} is used to perform GP regression along with the ROS Toolbox package.
} 

We used three tarps of blue, gray, and red colors as a proxy of regions with different intensities of toxicants. We collected our real-world data by flying the UAV at 3 different altitudes of 10, 20, and 30 meters in a boustrophedon pattern ($3.25\times4.73$, $6.51\times9.45$, $9.76\times14.12$ square meters, respectively, in the camera footprint size) in the environment of $30\times 30\times30$ cubic meters. We gathered 346 images spanned over these altitudes. 
Figure~\ref{fig:real_data} shows the ground truth. 
For each image in our dataset, we assigned the intensity value $f(\textbf{x})$ for each color on a per-pixel basis. We have three discrete intensity values of $3$, $2$, and $1$ associated with HSV color ranges of the red, gray, and blue tarps, respectively. 
The left image in Figure~\ref{fig:real_data} shows the visualization of the intensity values in Universal Transverse Mercator (UTM) coordinates. Darker regions represent a higher intensity value.

In the simulation, the original resolution of the image $240\times432$ was reduced to $19\times34$ resulting in $646$ pixel measurements.
\changed{Note that down-sampling is not required for GP computation, owing to the use of sparse GP, but required for fast data transmission.}
The camera footprint at an arm depends on the yaw of the UAV as well as the UAV position $\textbf{v}$. To use the intensity matrix with MF-GP-UCB, we do the mapping from image coordinates to real-world coordinates (\ie, UTM coordinates). To estimate UTM coordinates for each pixel, we make three main assumptions: (1) The principal point of the camera aligns with the image center. (2) Consequently, the UTM northing and easting of the UAV can be projected directly onto the image center. (3) Roll and pitch errors of the UAV are negligible.
\changed{To estimate world coordinates of each pixel position, we rotate the frame of reference to align it with the  UTM coordinate axis. Then, we apply the Pinhole camera model~\cite{hartley2003multiple} using the intrinsic parameters of the camera sensor obtained from camera calibration. }


\begin{figure}[htb]
\centering
\subfigure[The resultant GP mean.]{\includegraphics[width=0.49\columnwidth]{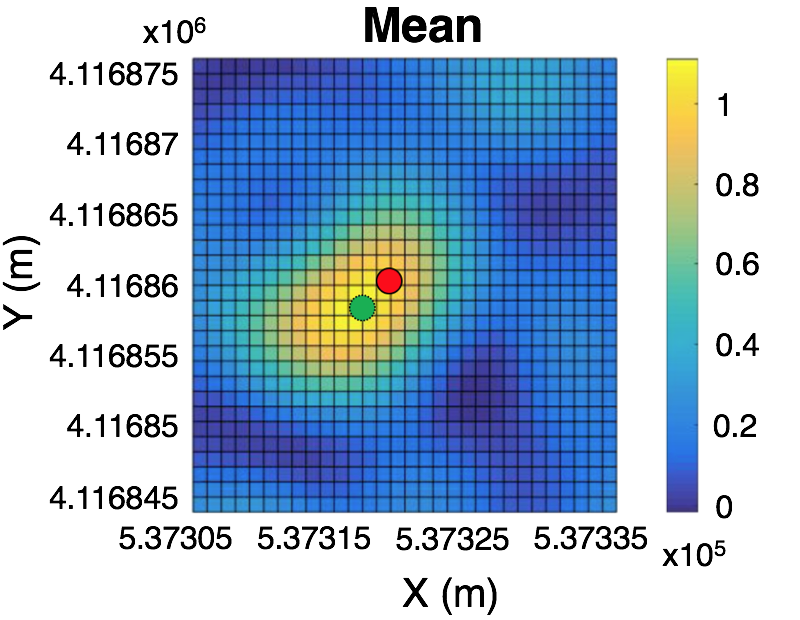}}
\subfigure[The resultant GP variance.]{\includegraphics[width=0.49\columnwidth]{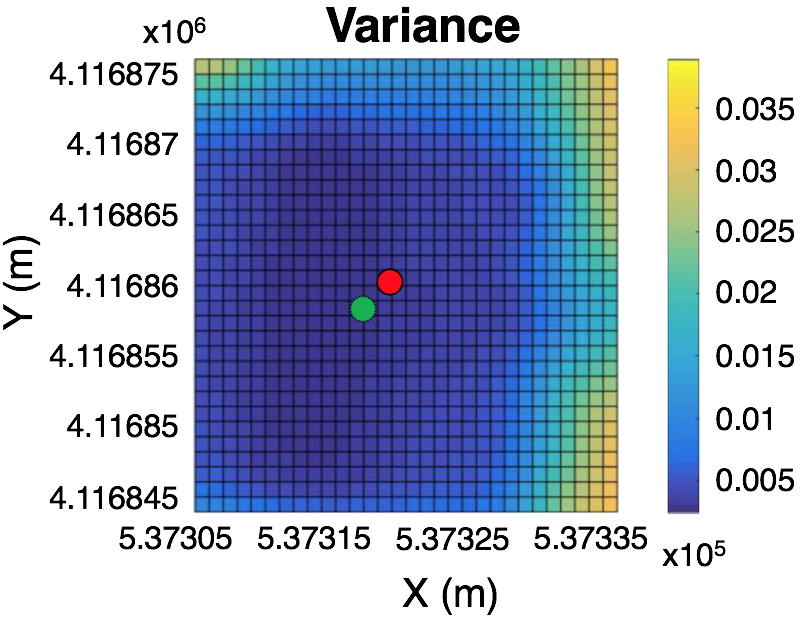}}
\caption{Simulation results using real-world data. The red solid circle and the green dotted circle denote $\textbf{x}^{OPT}$ and $\textbf{x}^{ALG}$, respectively.}
\label{fig:result}
\end{figure}

Figure~\ref{fig:result} shows the terminal states of GP by applying the proposed algorithm to real-world data. $\textbf{x}^{OPT}$ is $(537318m,4116861m)$ and $\textbf{x}^{ALG}$ is $(537316m,4116859m)$ in UTM coordinates. The point performance metric we obtained is $100\%$ as both $f(\textbf{x}^{OPT})$ and $f(\textbf{x}^{ALG})$ are $3$ while the arm performance metric is $66.7\%$.


\changed{
\subsection{Large-Scale Simulated Experiments}\label{subsec:large-scale}
To evaluate the scalability of our algorithm in a large-scale environment, we used the Gazebo simulator (Figure~\ref{fig:gazebo}). We used the hector\_quadrotor~\cite{kinetic} package in Gazebo to simulate the UAV model. To emulate a real-world scenario, an image of oil spill in San Francisco bay~\cite{inaglory_2007} is used as the ground plane in Gazebo to model the plume spread over an area of $200\times200$ square meters. We randomly chose five initial locations of the UAV within the environment and compared the performance with baselines (\ie, Boustrophedon, Gradient ascent, and variance reduction). The three  altitudes were set to be $10$, $15$, and $20$ meters. The number of arms at each altitude was $400$. 

\begin{figure}[htb]
\centering
{\includegraphics[width=0.56\columnwidth]{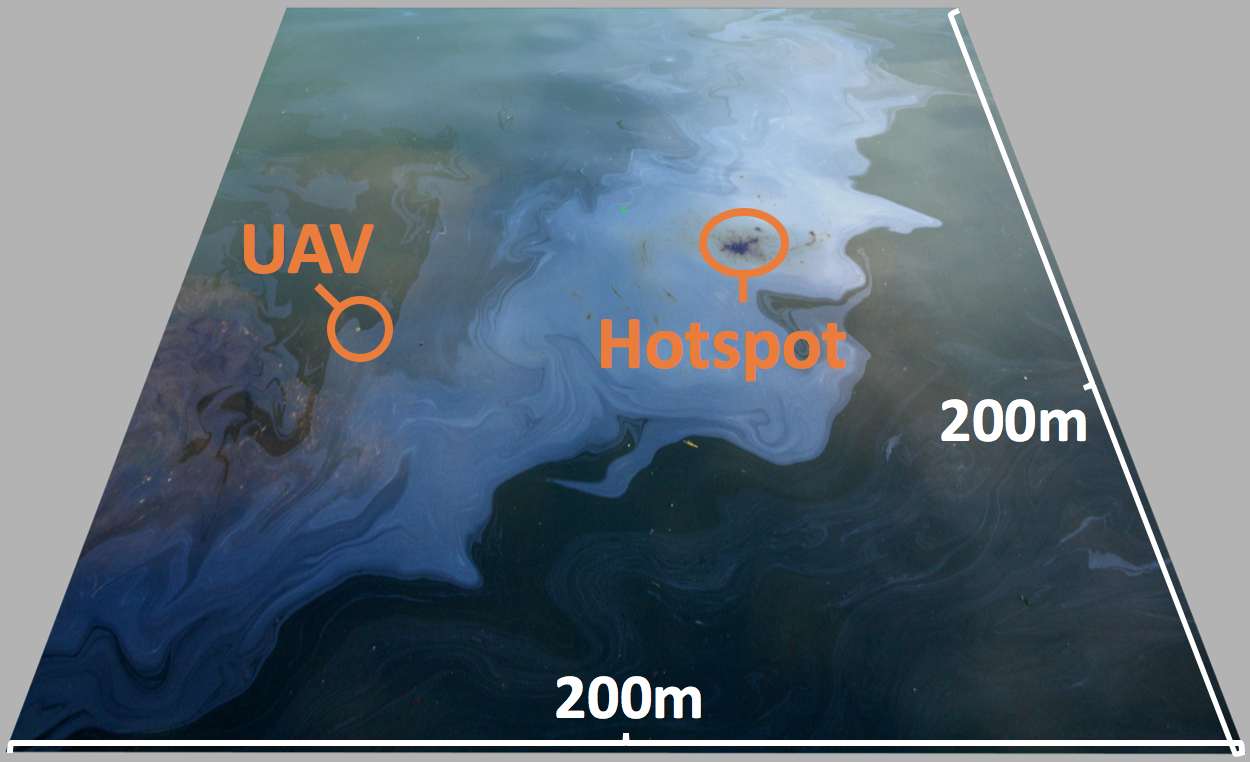}}
{\includegraphics[width=0.42\columnwidth]{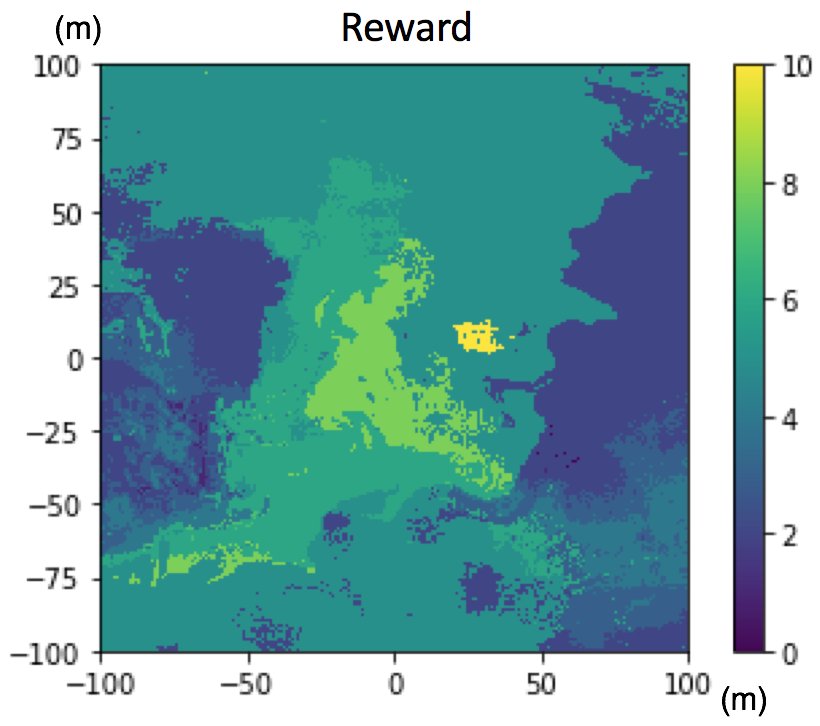}}
\caption{Image on the left shows a snapshot of the Gazebo simulation environment. The image of oil spill in San Francisco is used as ground plane in Gazebo to model the plume. Image on the right shows the Ground-truth reward map for the plume.}
\label{fig:gazebo}
\end{figure}

We computed the average performance values over both point and arm performance metrics from all five runs when the budget of $5000$ was given.
The average performance values we obtained are $83.05\pm 3.16\%$ for MF-GP-UCB, $64.14\pm 2.84\%$ for Boustrophedon, $32.86\pm 3.34\%$ for Gradient ascent, and $70.62\pm 5.00\%$ for variance reduction. We also observed that if a relatively small amount of budget is given, MF-GP-UCB (as well as any strategies) might not succeed as the robot does not explore the enough environment. This result supports our comparison analysis result as long as a certain amount of environment exploration is guaranteed by assigning a reasonable budget.

}

\section{Conclusion}\label{sec:conc}

In this paper, we propose an exploration algorithm that can find a hotspot in an unknown environment in limited time. We show how the UAV equipped with a downward-facing camera can be adopted to MAB and present simulation results to verify the performance of the algorithm. In particular, our proposed algorithm outperforms numerous baselines. Our empirical results indicate that incorporating CPV results in almost $10\%$ improvement in performance and sparse GPs results in orders of magnitude lower computational time without sacrificing any performance.

Immediate future work would be to conduct more rigorous real-world experiments using the proposed scheme. 
Analyzing the regret bound or suboptimality with respect to the optimal reward would be promising.

\bibliographystyle{IEEEtran}
\bibliography{IEEEabrv,yoon_refs}

\begin{thebibliography}{10}
\providecommand{\url}[1]{#1}
\csname url@samestyle\endcsname
\providecommand{\newblock}{\relax}
\providecommand{\bibinfo}[2]{#2}
\providecommand{\BIBentrySTDinterwordspacing}{\spaceskip=0pt\relax}
\providecommand{\BIBentryALTinterwordstretchfactor}{4}
\providecommand{\BIBentryALTinterwordspacing}{\spaceskip=\fontdimen2\font plus
\BIBentryALTinterwordstretchfactor\fontdimen3\font minus
  \fontdimen4\font\relax}
\providecommand{\BIBforeignlanguage}[2]{{%
\expandafter\ifx\csname l@#1\endcsname\relax
\typeout{** WARNING: IEEEtran.bst: No hyphenation pattern has been}%
\typeout{** loaded for the language `#1'. Using the pattern for}%
\typeout{** the default language instead.}%
\else
\language=\csname l@#1\endcsname
\fi
#2}}
\providecommand{\BIBdecl}{\relax}
\BIBdecl

\bibitem{sung2019competitive}
Y.~Sung and P.~Tokekar, ``A competitive algorithm for online multi-robot
  exploration of a translating plume,'' in \emph{2019 International Conference
  on Robotics and Automation (ICRA)}.\hskip 1em plus 0.5em minus 0.4em\relax
  IEEE, 2019, pp. 3391--3397.

\bibitem{nsf}
\BIBentryALTinterwordspacing
``Nri: Coordinated detection and tracking of hazardous agents with aerial and
  aquatic robots to inform emergency responders,'' Oct 2016. [Online].
  Available: \url{https://nsf.gov/awardsearch/showAward?AWD_ID=1637915}
\BIBentrySTDinterwordspacing

\bibitem{powers2018tracking}
C.~Powers, R.~Hanlon, and D.~Schmale, ``Tracking of a fluorescent dye in a
  freshwater lake with an unmanned surface vehicle and an unmanned aircraft
  system,'' \emph{Remote Sensing}, vol.~10, no.~1, p.~81, 2018.

\bibitem{tokekar2016sensor}
P.~Tokekar, J.~Vander~Hook, D.~Mulla, and V.~Isler, ``Sensor planning for a
  symbiotic uav and ugv system for precision agriculture,'' \emph{IEEE
  Transactions on Robotics}, vol.~32, no.~6, pp. 1498--1511, 2016.

\bibitem{cliff2015online}
O.~M. Cliff, R.~Fitch, S.~Sukkarieh, D.~L. Saunders, and R.~Heinsohn, ``Online
  localization of radio-tagged wildlife with an autonomous aerial robot
  system,'' in \emph{Robotics: Science and Systems}, 2015.

\bibitem{srinivas2010gaussian}
N.~Srinivas, A.~Krause, S.~Kakade, and M.~Seeger, ``Gaussian process
  optimization in the bandit setting: No regret and experimental design,'' in
  \emph{Proceedings of the 27th International Conference on Machine Learning},
  no. CONF.\hskip 1em plus 0.5em minus 0.4em\relax Omnipress, 2010.

\bibitem{quinonero2005unifying}
J.~Qui{\~n}onero-Candela and C.~E. Rasmussen, ``A unifying view of sparse
  approximate gaussian process regression,'' \emph{Journal of Machine Learning
  Research}, vol.~6, no. Dec, pp. 1939--1959, 2005.

\bibitem{titsias2009variational}
M.~Titsias, ``Variational learning of inducing variables in sparse gaussian
  processes,'' in \emph{Artificial Intelligence and Statistics}, 2009, pp.
  567--574.

\bibitem{dunbabin2012robots}
M.~Dunbabin and L.~Marques, ``Robots for environmental monitoring: Significant
  advancements and applications,'' \emph{IEEE Robotics \& Automation Magazine},
  vol.~19, no.~1, pp. 24--39, 2012.

\bibitem{lan2016rapidly}
X.~Lan and M.~Schwager, ``Rapidly exploring random cycles: Persistent
  estimation of spatiotemporal fields with multiple sensing robots,''
  \emph{IEEE Transactions on Robotics}, vol.~32, no.~5, pp. 1230--1244, 2016.

\bibitem{low2012decentralized}
K.~H. Low, J.~Chen, J.~M. Dolan, S.~Chien, and D.~R. Thompson, ``Decentralized
  active robotic exploration and mapping for probabilistic field classification
  in environmental sensing,'' in \emph{Proceedings of the 11th International
  Conference on Autonomous Agents and Multiagent Systems-Volume 1}.\hskip 1em
  plus 0.5em minus 0.4em\relax Citeseer, 2012, pp. 105--112.

\bibitem{hajieghrary2016multi}
H.~Hajieghrary, M.~A. Hsieh, and I.~B. Schwartz, ``Multi-agent search for
  source localization in a turbulent medium,'' \emph{Physics Letters A}, vol.
  380, no.~20, pp. 1698--1705, 2016.

\bibitem{silic2019field}
M.~Silic and K.~Mohseni, ``Field deployment of a plume monitoring uav flock,''
  \emph{IEEE Robotics and Automation Letters}, vol.~4, no.~2, pp. 769--775,
  2019.

\bibitem{galceran2013survey}
E.~Galceran and M.~Carreras, ``A survey on coverage path planning for
  robotics,'' \emph{Robotics and Autonomous systems}, vol.~61, no.~12, pp.
  1258--1276, 2013.

\bibitem{choset2000coverage}
H.~Choset, ``Coverage of known spaces: The boustrophedon cellular
  decomposition,'' \emph{Autonomous Robots}, vol.~9, no.~3, pp. 247--253, 2000.

\bibitem{flaspohler2018near}
G.~Flaspohler, N.~Roy, and Y.~Girdhar, ``Near-optimal irrevocable sample
  selection for periodic data streams with applications to marine robotics,''
  in \emph{2018 IEEE International Conference on Robotics and Automation
  (ICRA)}.\hskip 1em plus 0.5em minus 0.4em\relax IEEE, 2018, pp. 1--8.

\bibitem{manjanna2018heterogeneous}
S.~Manjanna, A.~Q. Li, R.~N. Smith, I.~Rekleitis, and G.~Dudek, ``Heterogeneous
  multi-robot system for exploration and strategic water sampling,'' in
  \emph{2018 IEEE International Conference on Robotics and Automation
  (ICRA)}.\hskip 1em plus 0.5em minus 0.4em\relax IEEE, 2018, pp. 1--8.

\bibitem{ma2018multi}
K.-C. Ma, Z.~Ma, L.~Liu, and G.~S. Sukhatme, ``Multi-robot informative and
  adaptive planning for persistent environmental monitoring,'' in
  \emph{Distributed Autonomous Robotic Systems}.\hskip 1em plus 0.5em minus
  0.4em\relax Springer, 2018, pp. 285--298.

\bibitem{hollinger2014sampling}
G.~A. Hollinger and G.~S. Sukhatme, ``Sampling-based robotic information
  gathering algorithms,'' \emph{The International Journal of Robotics
  Research}, vol.~33, no.~9, pp. 1271--1287, 2014.

\bibitem{binney2013optimizing}
J.~Binney, A.~Krause, and G.~S. Sukhatme, ``Optimizing waypoints for monitoring
  spatiotemporal phenomena,'' \emph{The International Journal of Robotics
  Research}, vol.~32, no.~8, pp. 873--888, 2013.

\bibitem{snoek2012practical}
J.~Snoek, H.~Larochelle, and R.~P. Adams, ``Practical bayesian optimization of
  machine learning algorithms,'' in \emph{Advances in neural information
  processing systems}, 2012, pp. 2951--2959.

\bibitem{slivkins2019introduction}
A.~Slivkins \emph{et~al.}, ``Introduction to multi-armed bandits,''
  \emph{Foundations and Trends{\textregistered} in Machine Learning}, vol.~12,
  no. 1-2, pp. 1--286, 2019.

\bibitem{reverdy2014modeling}
P.~B. Reverdy, V.~Srivastava, and N.~E. Leonard, ``Modeling human decision
  making in generalized gaussian multiarmed bandits,'' \emph{Proceedings of the
  IEEE}, vol. 102, no.~4, pp. 544--571, 2014.

\bibitem{guha2009multi}
S.~Guha and K.~Munagala, ``Multi-armed bandits with metric switching costs,''
  in \emph{International Colloquium on Automata, Languages, and
  Programming}.\hskip 1em plus 0.5em minus 0.4em\relax Springer, 2009, pp.
  496--507.

\bibitem{vansteenwegen2011orienteering}
P.~Vansteenwegen, W.~Souffriau, and D.~Van~Oudheusden, ``The orienteering
  problem: A survey,'' \emph{European Journal of Operational Research}, vol.
  209, no.~1, pp. 1--10, 2011.

\bibitem{audibert2010best}
J.-Y. Audibert and S.~Bubeck, ``Best arm identification in multi-armed
  bandits,'' 2010.

\bibitem{kandasamy2019multi}
K.~Kandasamy, G.~Dasarathy, J.~Oliva, J.~Schneider, and B.~Poczos,
  ``Multi-fidelity gaussian process bandit optimisation,'' \emph{Journal of
  Artificial Intelligence Research}, vol.~66, pp. 151--196, 2019.

\bibitem{song2019general}
J.~Song, Y.~Chen, and Y.~Yue, ``A general framework for multi-fidelity bayesian
  optimization with gaussian processes,'' in \emph{The 22nd International
  Conference on Artificial Intelligence and Statistics}, 2019, pp. 3158--3167.

\bibitem{wei2020expedited}
L.~Wei, X.~Tan, and V.~Srivastava, ``Expedited multi-target search with
  guaranteed performance via multi-fidelity gaussian processes,'' 2020.

\bibitem{hartley2003multiple}
R.~Hartley and A.~Zisserman, \emph{Multiple view geometry in computer
  vision}.\hskip 1em plus 0.5em minus 0.4em\relax Cambridge university press,
  2003.

\bibitem{rasmussen2003gaussian}
C.~E. Rasmussen, ``Gaussian processes in machine learning,'' in \emph{Summer
  School on Machine Learning}.\hskip 1em plus 0.5em minus 0.4em\relax Springer,
  2003, pp. 63--71.

\bibitem{krause2010sfo}
A.~Krause, ``Sfo: A toolbox for submodular function optimization,'' \emph{The
  Journal of Machine Learning Research}, vol.~11, pp. 1141--1144, 2010.

\bibitem{quigley2009ros}
M.~Quigley, K.~Conley, B.~Gerkey, J.~Faust, T.~Foote, J.~Leibs, R.~Wheeler, and
  A.~Y. Ng, ``Ros: an open-source robot operating system,'' in \emph{ICRA
  workshop on open source software}, vol.~3, no. 3.2.\hskip 1em plus 0.5em
  minus 0.4em\relax Kobe, Japan, 2009, p.~5.

\bibitem{kinetic}
\BIBentryALTinterwordspacing
ROS-hector\_quadrotor, ``Wiki.'' [Online]. Available:
  \url{http://wiki.ros.org/hector_quadrotor}
\BIBentrySTDinterwordspacing

\bibitem{inaglory_2007}
\BIBentryALTinterwordspacing
B.~Inaglory, ``Oil spill in san francisco bay.jpg,'' Dec 2007. [Online].
  Available:
  \url{https://commons.wikimedia.org/wiki/File:Oil_spill_in_San_Francisco_bay.jpg}
\BIBentrySTDinterwordspacing

\end{thebibliography}


\end{document}